\def\paperID{9046}
\def\confName{CVPR}
\def\confYear{2024}

\def\paperTitle{M$^4$-BLIP: Advancing Multi-Modal Media Manipulation Detection through Face-Enhanced Local Analysis} 

\def\authorBlock{
Hang~Wu,~
Ke~Sun,~
Jiayi~Ji,~
Xiaoshuai~Sun,~
Rongrong~Ji,~\\
Key Laboratory of Multimedia Trusted Perception and Efficient Computing \\
}

\newif\ifreview 
\newif\ifarxiv \newcommand{\arxiv}{\arxivtrue}
\newif\ifcamera 
\newif\ifrebuttal 

\arxiv 

\pdfoutput=1
\documentclass[10pt,twocolumn,letterpaper]{article}
\ifreview \usepackage[review]{cvpr} \fi
\ifarxiv \usepackage[pagenumbers]{cvpr} \fi
\ifrebuttal \usepackage[rebuttal]{cvpr} \fi
\ifcamera \usepackage{cvpr} \fi

\usepackage{graphicx}	
\usepackage{amsmath}	
\usepackage{amssymb}	
\usepackage{booktabs}
\usepackage{times}
\usepackage{microtype}
\usepackage{epsfig}
\usepackage{caption}
\usepackage{float}
\usepackage{placeins}
\usepackage{color, colortbl}
\usepackage{stfloats}
\usepackage{enumitem}
\usepackage{tabularx}
\usepackage{xstring}
\usepackage{multirow}
\usepackage{xspace}
\usepackage{url}
\usepackage{subcaption}
\usepackage{xcolor}
\usepackage[hang,flushmargin]{footmisc}

\ifcamera \usepackage[accsupp]{axessibility} \fi

% ...

\ifarxiv  \fi

\newcommand{\R}[1]{{%
    \textbf{%
        \ifstrequal{#1}{1}{\textcolor{red}{R#1}}{%
        \ifstrequal{#1}{2}{\textcolor{blue}{R#1}}{%
        \ifstrequal{#1}{3}{\textcolor{magenta}{R#1}}{%
        \ifstrequal{#1}{4}{\textcolor{teal}{R#1}}{%
                           \textcolor{cyan}{R#1}%
        }}}}%
    }%
}}

\usepackage{xr-hyper}

\makeatletter
\newcommand*{\addFileDependency}[1]{
  \typeout{(#1)}
  \@addtofilelist{#1}
  \IfFileExists{#1}{}{\typeout{No file #1.}}
}

\makeatother
\newcommand*{\myexternaldocument}[1]{
    \externaldocument{#1}
    \addFileDependency{#1.tex}
    \addFileDependency{#1.aux}
}

\definecolor{cvprblue}{rgb}{0.21,0.49,0.74}
\usepackage[pagebackref,breaklinks,colorlinks,citecolor=cvprblue]{hyperref}
\usepackage[capitalize]{cleveref}
\crefname{section}{Sec.}{Secs.}
\crefname{table}{Table}{Tables}
\crefname{figure}{Fig.}{Figs.}

\unless\ifarxiv \myexternaldocument{_supplementary} \fi

\begin{document}
\title{\paperTitle}
\author{\authorBlock}
\maketitle

\begin{abstract}
In the contemporary digital landscape, multi-modal media manipulation has emerged as a significant societal threat, impacting the reliability and integrity of information dissemination. Current detection methodologies in this domain often overlook the crucial aspect of localized information, despite the fact that manipulations frequently occur in specific areas, particularly in facial regions. In response to this critical observation, we propose the M$^4$-BLIP framework. This innovative framework utilizes the BLIP-2 model, renowned for its ability to extract local features, as the cornerstone for feature extraction. Complementing this, we incorporate local facial information as prior knowledge. A specially designed alignment and fusion module within M$^4$-BLIP meticulously integrates these local and global features, creating a harmonious blend that enhances detection accuracy. Furthermore, our approach seamlessly integrates with Large Language Models (LLM), significantly improving the interpretability of the detection outcomes. Extensive quantitative and visualization experiments validate the effectiveness of our framework against the state-of-the-art competitors.
\end{abstract}
\section{Introduction}

With the advancement of technology, multi-modal news data, represented by image-text pairs, has become increasingly prevalent in our lives. This proliferation, however, has given rise to a growing incidence of fabricated news, posing serious security risks. Face Forgery Detection plays a pivotal role in safeguarding information security by effectively identifying manipulated facial images and preventing the proliferation of deceptive content. Traditional face forgery detection methods have primarily focused on the recognition of forged images, achieving commendable results in the image modality. However, these methods have lacked attention to the textual modality. Fake news detection (FND) is dedicated to detecting fake information in text news\cite{conroy2015automatic}, which can help readers identify the bias and misinformation and eliminate the negative spreading. Similarly, it also ignores the importance of image modality for forgery detection. Thus, the development of multi-modal forgery detection models plays a crucial role in safeguarding information security.


\begin{figure}
    \centering
    \includegraphics[width=1\linewidth]{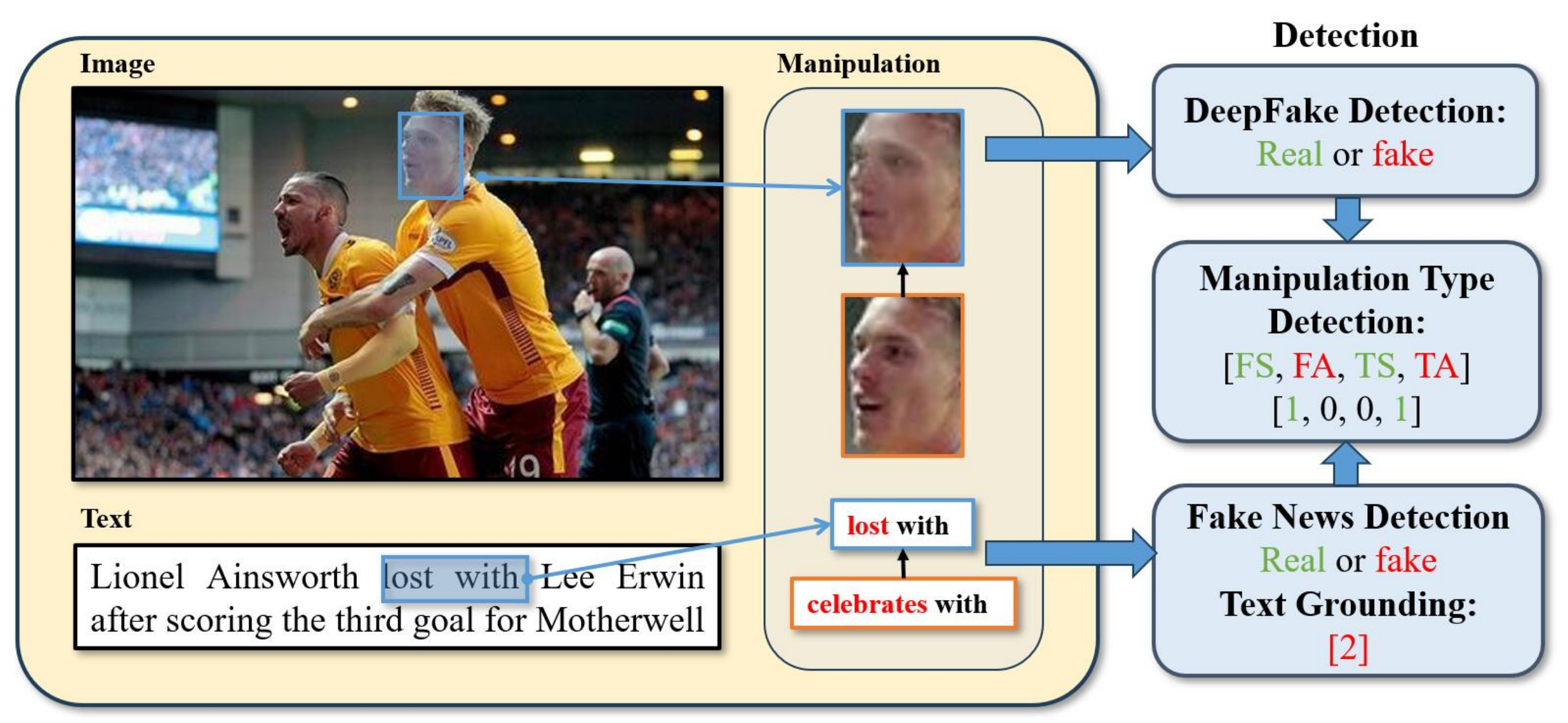}
    \caption{Examples of multi-modal manipulation detection (FS:Face Swap Manipulation, FA:aspect Attribute Manipulation, TS:Text Swap Manipulation, TA:Text Attribute Manipulation)}
    \label{fig:Paradigm}
\end{figure}


In recent years, many works have been conducted to detect anomalies in news and posts by aggregating multi-modal features~\cite{chen2022cross,khattar2019mvae,wang2018eann}. For example, SpotFake~\cite{singhal2019spotfake} only concat the image feature form VGG-19~\cite{simonyan2014very} and the text feature from BERT~\cite{devlin2018bert}to aggregating multi-modal features, SAFE~\cite{Zhou_Wu_Zafarani_2020} aggregate features by computing the similarity of image features and text features.
Traditional news detection methods typically address mismatches between images and texts, rather than manipulations within the images or texts themselves.
However, with advancements in the Deepfake method and large language models, there has been an increased focus on localized manipulation of faces and text. DGM$^4$~\cite{shao2023detecting} first introduced a new dataset specifically for this task and proposed a new paradigm for multi-modal manipulation detection tasks. As shown in Fig.~\ref{fig:Paradigm}, multi-modal manipulation detection aims to detect manipulations in text-image pairs, enabling the verification of the authenticity of images and text, classification of manipulation types, and localization of manipulated text. DGM$^4$ aims not only to detect the authenticity of multi-modal media but also to ground the manipulated content (i.e., text tokens manipulation), requiring deeper reasoning of multi-modal information.

In this paper, aligned with DGM$^4$~\cite{shao2023detecting}, we concentrate on multi-modal media manipulation detection tasks. Reevaluating this domain, we recognize that both image and text manipulations essentially represent localized alterations, highlighting the significance of local information. Consequently, our detection paradigm is designed with a focus on extracting and incorporating local features. This approach is two-fold: Firstly, the foundation model must possess the capability for local feature extraction. We observe that BLIP2~\cite{li2023blip}, due to its inclusion of fine-grained perception tasks like RES and REC during pre-training, exhibits stronger local information perception compared to global knowledge-oriented models like CLIP~\cite{radford2021learning}. Secondly, we aim to reinforce the model's capacity for local feature extraction and analysis by introducing facial local priors and enhancing self-knowledge through the fusion of local and global features.

Based on this design, we propose a local prior-enhanced multi-modal forgery detection framework built upon the BLIP-2 architecture, named M$^4$-BLIP. Specifically, we employ BLIP-2 pre-trained image and text encoders for initial feature extraction. Simultaneously, we utilize a specialized facial extractor to identify and extract local facial features, establishing them as crucial local priors. These local priors are instrumental in detecting subtle manipulations typically missed by global analysis.
Such local and global elements undergo feature alignment and fusion strategy, crafting a comprehensive understanding of both micro-details and wider contextual information. Concretely, the Fine-grained Contrastive Alignment module aligns global image-text features by bringing authentic pairs closer together while distancing the corresponding forged features. Subsequently, the Multi-modal Local-and-Global Fusion module, utilizes the Q-Former to fusion the global and local features with textual features independently and fusion with cross-attention, allowing for an insightful interplay between the global and local information.
Training employs specific loss functions for binary classification, manipulation type classification, and text manipulation localization, refining the model's precision. Furthermore, integration with large language models broadens M$^4$-BLIP's interpretability and practical utility in complex multi-modal forgery scenarios.

The primary contributions of this work can be summarized as follows:

\begin{itemize}
\item We are the first to introduce BLIP-2, a pre-trained model capable of extracting local features, along with local feature priors into the Multi-Modal Media Manipulation Detection task, which offers a new paradigm for the field.
\item We propose a novel multi-modal media forgery detection framework named M$^4$-BLIP. This framework enhances forgery detection capabilities by effectively aligning and fusing global and local features, ensuring more comprehensive feature extraction.
\item Our method integrates seamlessly with large language models, not only improving interpretability but also laying a foundational research basis for future fake news detection with large language models.
\item Extensive quantitative and visualization experiments validate the effectiveness of our framework against the state-of-the-art competitors.
\end{itemize}


\label{sec:intro}

\section{Related Work}
\subsection{Multimodal Forgery Detection}

Much of the prior multimodal forgery detection research has revolved around aligning multi-modal information. For instance, FND-CLIP\cite{zhou2023multimodal} incorporates the CLIP\cite{radford2021learning} model into the FND task. It harnesses CLIP's semantic understanding capabilities to assess the relationship between image and text, subsequently utilizing this correlation to compute fusion features. The classification results are then derived from these fused features.Similarly, CAFE\cite{chen2022cross} employs an adaptive approach to amalgamate multi-modal features and single-modal features by calculating their correlations. This adaptability allows for the selective aggregation of information, enhancing the discriminative power of the features.GAME-ON\cite{dhawan2022game} takes a novel approach by utilizing graph neural networks to facilitate deeper interactions between image and text features, consequently enabling the model to acquire more potent feature representations.In contrast, LIIMR\cite{singhal2022leveraging} prioritizes interpretability in decision-making by attenuating the influence of weaker modalities while accentuating the contributions of the stronger ones. Additionally, COOLANT\cite{wang2023cross} employs auxiliary tasks to tailor the contrastive learning process to the specific requirements of the FND task, enhancing its suitability and effectiveness. However, their attention has primarily been confined to scrutinizing the authenticity of image-text pairs, with limited exploration of other facets. Realizing this, DGM$4$\cite{shao2023detecting} introduces a novel paradigm in the realm of multi-modal detection tasks, aiming not only to detect the authenticity of the image-text pairs but also to detect the forged content (i.e., text tokens manipulation), DGM$4$ achieves this by aligning image and text embeddings, facilitated by operation-aware contrastive learning between two single-modal encoders. Furthermore, it assembles multi-modal embeddings through a profound operational rationale, employing modality-aware cross-attention within a multimodal aggregator. Notably, various detection heads are employed to address distinct tasks. However, it is crucial to recognize that these methods may not account for potential inconsistencies between the manipulated image-text and the original counterpart. We posit that addressing these inconsistencies can provide the model with additional information.

\subsection{Multimodal Representation Learning}

Multimodal representation learning, particularly visual-language representation learning, has garnered significant attention in recent years. The overarching objective of multimodal representation learning is to handle information originating from diverse modalities by bridging the gap that separates them. A prime example of this is CLIP\cite{radford2021learning}, which serves as a pivotal link between image and text. It was trained on a diverse array of image-text pairs, with the goal of predicting the most contextually relevant text snippet for a given image, without direct optimization for any specific task. As a result, CLIP finds versatile application in numerous downstream tasks related to both graphics and text, including the tackling of relatively straightforward challenges in multimodal forgery detection. More recently, visual language models, exemplified by models like BLIP-2\cite{li2023blip}, have made remarkable strides in understanding and generating multimodal content across open domains. Achieved through pre-training with visual language capabilities and the integration of large language models, these innovations underscore the potent capabilities of visual language pre-training models across a broad spectrum of downstream tasks. Inspired by these advancements, we transfer our multi-modal forgery detection model to  pre-trained visual language model. Which can imbues our model with enhanced multi-modal information comprehension, while also conferring upon it the ability to generate content, further extending its utility and efficacy.

\label{sec:related}

\section{Method}

\begin{figure*}
    \centering
    \includegraphics[width=0.95\linewidth]{figs/model.pdf}
    \caption{Overview of proposed methods. Different levels of image features are extracted through a global image encoder and a local deepfake detector. Subsequently, Q-Former is employed to aggregate these features from various levels along with text features. Finally, tasks are executed in a layered manner using different detection heads, combined with LLM, to generate task-related descriptions. (Best viewed in color.)}
    \label{fig:Overview}
\end{figure*}
Our research focused on detecting and classifying multi-modal forgeries, presents an overarching framework M$^4$-BLIP. As depicted in the Fig.~\ref{fig:Overview}, the framework starts with the extraction of both global and local visual features along with textual features using multi-modal pre-trained models and pre-trained deepfake model, followed by aligning these features via Fine-grained Contrastive Alignment(FCA). The Multi-modal Locol-and-Global Fusion(MLGF) module is then employed to fuse these features, enhancing the model's understanding of both details and context. The fused features are utilized for binary classification, multi-modal manipulation classification, and text manipulation detection tasks. Additionally, the integration of the model with large language models elevates its interpretability and applicability in multi-modal tasks.

\subsection{Feature Extraction and Representation}
This section examines the details of multi-modal feature extraction and representation. The process is divided into two parts: global feature extraction and local feature extraction.

In terms of global feature extraction, we utilize image and text encoders from multi-modal pre-trained models, designated as $E_v$ and $E_t$. These encoders extract comprehensive features from the original images $I$ and text $T$. The primary reason for employing pre-trained models is their proven ability to learn extensive semantic information from large datasets, thereby significantly enhancing our model's capacity to understand image content and analyze the relationship between images and text.

Focusing on local features, special attention is given to facial regions as forgeries like deepfake typically involve subtle manipulations in these areas. To capture these details accurately, a dedicated facial detection tool is used to extract facial regions from the original images $I$, denoted as $I^{'}$. These facial images are then processed by a specialized deepfake detector $E_d$. The role of $E_d$ is to conduct binary classification on the input facial images to ascertain the presence of forgery traces. This approach to local feature extraction enables our model to detect subtle changes induced by deepfake technology more sensitively, thus increasing the accuracy in detecting forged content. By integrating local features $e^d = E_d(I^{'})$ with global features $e^v = E_v(I)$ and text features $e^t = E_t(T)$, we provide a comprehensive perspective for the task of multi-modal forgery detection.

\label{sec:feature_extraction}

\subsection{Feature Alignment and Fusion}
After obtaining global/local visual features and textual features, to ensure their semantic coherence and to fully utilize their potential in forgery detection tasks, we adopt two key strategies: Fine-grained Contrastive Alignment (FCA) and Multi-modal Locol-and-Global Fusion (MLGF). The purpose of FCA is to ensure semantic consistency between images and texts, which is crucial for distinguishing authentic from forged image-text pairs. MLGF, on the other hand, aims to make global features interact and complement with local features, enhancing the model's sensitivity to details and improving accuracy in understanding the overall scene. The specifics of these two strategies are detailed as follows:

\noindent\textbf{Fine-grained Contrastive Alignment. }
For feature alignment, we leverage the contrastive learning paradigm. This approach aims to enhance semantic alignment within image-text pairs and assist the model in differentiating between authentic and forged pairs. Uniquely, our method not only tightens the embeddings of genuine image-text pairs but also separates the embeddings of original and forged text, as well as original and forged images. Such fine-grained contrastive ensures semantic congruency among authentic pairs while introducing distinctiveness at the embedding level for forgeries.

The process is represented by the following equations:
\begin{equation}
\mathcal{L}_{v2t} = -\mathbb{E}_{p(I, T)}\left[\log \frac{\exp \left(S\left(I^{+}, T^{+}\right) / \tau\right)}{\sum_{k=1}^K \exp \left(S\left(I^{+}, T_k^{-}\right) / \tau\right)}\right]  
\end{equation}
\begin{equation}
\mathcal{L}_{t2v} = -\mathbb{E}_{p(I, T)}\left[\log \frac{\exp \left(S\left(T^{+}, I^{+}\right) / \tau\right)}{\sum_{k=1}^K \exp \left(S\left(T^{+}, I_k^{-}\right) / \tau\right)}\right]   
\end{equation}
$I^{+}$, $T^{+}$, $I^{-}$, $T^{-}$ are sets of positive and negative samples from image-text pairs, where $I^{-}$, $T^{-}$ are the manipulation samples corresponding of $I^{+}$, $T^{+}$, and $S(I, T) = h_v\left(e^v\right)^{\mathrm{T}} h_t\left(e^t\right)$ with $h_v$, $h_t$ being two prediction heads that map embeddings to the same dimension. Thus, the loss function for the Cross-modal Alignment module is:
\begin{equation}
\mathcal{L}_{ITC} = (\mathcal{L}_{v2t} + \mathcal{L}_{t2v}) / 2
\end{equation}
Such an alignment method not only emphasizes accurate recognition of authentic content but also provides additional discernibility in handling forgeries. The inclusion of fine-grained negative samples enhances the model's perception of discontinuous image-text pairs, thereby contributing to improved generalization capabilities.

\label{sec:Feature_Alignment_and_Fusion}
\noindent\textbf{Multi-modal Locol-and-Global Fusion. }
To effectively integrate local and global image features with textual features, inspired by BLIP2, we utilize the Q-Former structure \(Q\) for the fusion of visual and textual features. Q-Former consists of learnable query embeddings and two transformer modules sharing self-attention layers, one interacting with image features and the other with text features. Query embeddings engage in mutual interactions through self-attention layers and with image features through cross-attention layers, capable of interacting with textual features due to shared self-attention layers. This selective fusion mechanism aids in learning informative aspects from both image and text modalities, contributing to subsequent tasks.

To better integrate local and global information, we separately interact text \(T\) with global image embedding \(e^v\) and local embedding \(e^d\), resulting in \(f^v=Q(e^v, T)\) and \(f^d=Q(e^d, T)\). The motivation behind this approach is to combine different scales of visual features with textual features, allowing our model to understand the minute details within images as well as grasp the overall context. Simultaneously, to ensure that the fused local features are more focused on binary classification, while the fused global features concentrate on manipulation type classification, we designed the following loss functions:
\begin{equation}
\mathcal{L}_{d} = \mathbb{E}_{(I, T) \sim P} \mathbf{H}\left(C_b\left(f^d\right), L\right)
\end{equation}
\begin{equation}
\mathcal{L}_{v} = \mathbb{E}_{(I, T) \sim P} \mathbf{H}\left(C_m\left(f^v\right), L_{\mathrm{mul}}\right)
\end{equation}
Here, \(\mathbb{E}\) denotes the expectation operation, calculating the average loss across different image-text pair sample sets \(P\). The function \(\mathbf{H}\) is the cross-entropy function, measuring the discrepancy between the model's predictions and the actual labels. \(C_b\) and \(C_m\) are multilayer perceptrons (MLPs) used for binary and multiclass classification, corresponding to local features \(f^d\) and global features \(f^v\), respectively. 
This meticulous fusion strategy significantly enhances the model's performance in complex multi-modal scenarios, ensuring accuracy in authenticity classification and manipulation type identification.

Finally, we employ a cross-attention module to fuse \(f^d\) and \(f^v\):
\begin{equation}
f = \operatorname{CrossAttention}(f^d, f^v)
\end{equation}
\begin{equation}
\operatorname{CrossAttention}(Q, K, V)=\operatorname{Softmax}\left(K^T Q / \sqrt{D}\right) V
\end{equation}
We utilize the fused feature \(f\) for the final loss function calculation. The global features provide macroscopic contextual information to the local features, which is crucial for the accurate detection of deepfake content~\cite{zhou2017two}. Meanwhile, the local features direct the global features to focus more on facial regions, providing finer-grained features of forgeries. In essence, through the fusion of local and global features, our model gains a comprehensive understanding and recognition of subtle changes within image-text pairs, thereby significantly enhancing the accuracy of forgery detection.

\subsection{Feature Classifier and Detector}
To more accurately identify multi-modal forgeries, we categorize our detector into three components: binary classification, manipulation type classification, and text manipulation localization. Leveraging the fused features \(f=\{f_{\mathrm{cls}}, f_{\mathrm{cls}}\}\) obtained through the cross-modal fusion module Q-Former, the [cls] token \(q_{\mathrm{cls}}\), after interacting with both textual and image features, comprehensively learns the authenticity status and manipulation type of the image-text pairs. Consequently, we utilize the [cls] token for real and fake detection and manipulation type classification. Our loss function is defined as follows:
\begin{equation}
\mathcal{L}_{MLC}=\mathbb{E}_{(I, T) \sim P} \mathbf{H}\left(C_m\left(f_{\mathrm{cls}}\right), L_{\mathrm{mul}}\right)
\end{equation}
\begin{equation}
\mathcal{L}_{BLC}=\mathbb{E}_{(I, T) \sim P} \mathbf{H}\left(C_b\left(f_{\mathrm{cls}}\right), L\right)
\end{equation}
Due to shared self-attention layers with the text encoder, the remaining tokens \(q_{\mathrm{tok}}\) in the query token comprehensively learn forged information within the text. Integrating information from the image facilitates the effective identification of regions in the text that do not align with the image information and exhibit inherent disharmony. Therefore, we utilize these tokens for text manipulation localization. The specific process is outlined as follows:
\begin{equation}
\mathcal{L}_{TMG}=\mathbb{E}_{(I, T) \sim P} \mathbf{H}\left(D_t\left(f_{\mathrm{tok}}\right), L_{\mathrm{tok}}\right)
\end{equation}
where \(D_t\) is a detector for text manipulation location detection, \(L_{\mathrm{tok}}\) is the label for manipulation localization.

\noindent\textbf{Overall Loss Function.} Considering both the
local-and-global loss and aforementioned loss, the overall loss for our proposed framework is:
\begin{equation}
    L = {L}_{TMG} + {L}_{BLC} + {L}_{MLC} + {L}_{v} + {L}_{d},
\end{equation}

\subsection{Integration with LLM}

To enhance the interpretability of our model and to expand its applicability in multi-modal tasks, we integrate the model with large language models (LLM). This integration not only endows the model with the capability to generate text but also allows it to express complex relationships between multi-modal data more intuitively. Specifically, we project the features obtained by the model through a simple fully connected layer \(h_{LLM}\) to the same dimension as the text input of the LLM. Subsequently, the projected features are combined with textual instruction and jointly input into the LLM. Leveraging the generative capacity of the LLM, our model can produce text relevant to the current task, thus enhancing the interpretability and user-friendliness of the results. The specific process is as follows:
\begin{equation}
output = \operatorname{LLM}\left(h_{LLM}(f) + \text{Instruction}\right)
\end{equation}
Following the approach of MiniGPT-4, we finetuned the fully connected layer \(h_{LLM}\) with the following template:

\textit{\#\#\#Human:(Img)(ImageFeature)(/Img)(Instruction) \#\#\#Assistant:}

Here, \textit{ImageFeature} in this template represents the fusion feature \(f\) and \textit{Instruction} represents different task instructions,  LLM will answer specified questions according to different instruction. Such as the instruction "Is this image-text pair manipulated?" denoting the binary classification task, "What type of manipulation is involved?" denoting the manipulation type classification task.
\label{sec:method}

\section{Experiments}
\subsection{Implementation Details}
\begin{table*}[!h]
    \renewcommand\arraystretch{1.1}
    \centering
    
    \begin{tabular}{c|ccc|ccc|ccc}
    \toprule
    Method & \multicolumn{3}{c|}{\textit{Binary Cls}} &\multicolumn{3}{c|}{\textit{Multi-Label Cls}} &\multicolumn{3}{c}{\textit{Text Grounding}}\\
    \hline
    Categories& AUC& EER &ACC& mAP& CF1& OF1& Precision& Recall& F1\\
    \hline
    CLIP\cite{radford2021learning}& 83.22& 24.61&76.40& 66.00& 59.52& 62.31& 58.12& 22.11& 32.03\\

    VILT\cite{kim2021vilt}& 85.16& 22.88&78.38& 72.37& 66.14& 66.00& 66.48& 49.88& 57.00\\

    ALBEF\cite{li2021align}& 86.95& 19.89&79.75& 84.53& 73.04& 74.32& 68.02& 61.08& 64.37\\

    BLIP-2\cite{li2023blip}& 89.96& 18.09&82.17& 83.63& 76.02& 76.37& 80.09& 59.39& 68.20\\

    HAMMER\cite{shao2023detecting}& 93.19& 14.10&86.39& 86.22& 79.37& 80.37& 75.01& 68.02& 71.35\\
   \hline
   \textbf{Ours}& \textbf{94.10}& \textbf{13.25}& \textbf{86.92}&\textbf{87.97}& \textbf{79.97}& \textbf{80.72}& \textbf{83.71}& \textbf{71.35}&\textbf{76.87}\\
   \bottomrule
    \end{tabular}
    \caption{Comparison with multi-modal learning methods in binary classification, multi-label classification and manipulation grounding task. Our method surpasses the current state-of-the-art methods and widely used multimodal learning approaches across all metrics. }
    \label{table:1}
\end{table*}

\begin{table}[!h]
    \centering
    \renewcommand\arraystretch{1.2}
    \resizebox{0.28\textwidth}{!}{
    \begin{tabular}{c|ccc}
    \toprule
    Method & \multicolumn{3}{c}{\textit{Binary Cls}} \\
    \hline
    Categories& AUC& EER &ACC\\
    \hline
    TS\cite{luo2021generalizing}& 91.80& 17.11&82.89\\
    MAT\cite{zhao2021multi}& 91.31& 17.65&82.36\\
   \hline
   \textbf{Ours}& \textbf{93.73}& \textbf{13.04}& \textbf{87.01}\\
   \bottomrule
    \end{tabular}
    }
    \caption{Comparison of image-modal methods. In the binary classification task for images, our multi-modal approach significantly outperforms the current uni-modal methods.}
    \label{table:2}
\end{table}
\begin{table}[!h]
    \centering
    \renewcommand\arraystretch{1.2}
    \resizebox{\columnwidth}{!}{
    
    \begin{tabular}{c|ccc|ccc}
    \toprule
    Method & \multicolumn{3}{c|}{\textit{Binary Cls}}& \multicolumn{3}{c}{\textit{Text Grounding}} \\
    \hline
    Categories& AUC& EER &ACC&Precision& Recall& F1\\
    \hline
    BERT\cite{devlin2018bert}& 80.82& 28.02&68.98&41.39&63.85&50.23\\
    LUKE\cite{yamada2020luke}& 81.39& 27.88&76.18&50.52&37.93&43.33\\
   \hline
   \textbf{Ours}& \textbf{93.72}& \textbf{13.46}& \textbf{86.29}& \textbf{72.80}& \textbf{65.10}&\textbf{68.74}\\
   \bottomrule
    \end{tabular}
    }
    \caption{Comparison of text-modal methods. In both the binary classification and the text manipulation grounding task, our multi-modal approach significantly outperforms current uni-modal methods.}
    \label{table:3}
\end{table}
\label{sec:experiments}
\textbf{Datasets.}
We utilized DGM$^4$\cite{shao2023detecting} as our dataset, which comprises various image-text pairs generated by different manipulation types. The dataset includes not only the authenticity information of corresponding image-text pairs and the associated manipulation types but also manipulation localization details such as the position of manipulated text. Leveraging this information allows the model to learn more about manipulation types and relevant localization details, thereby aiding in the detection of authenticity.

\textbf{Implement Details.}
To leverage visual-language correspondence information, we use BLIP-2\cite{li2023blip} as the pretraining model for Q-Former $Q$ and the text encoder $E_t$, and choose ViT-g/14 from EVA-CLIP\cite{fang2023eva} as the image encoder $E_v$. We resize the input image to 224 × 224, applying random resized cropping and horizontal flipping for augmentation. For the deepfake network $E_d$, we select EfficientNet-b4\cite{tan2019efficientnet} as the backbone. Following BLIP-2, we use the AdamW\cite{loshchilov2017decoupled} optimizer with $\beta_1=0.9$, $\beta_2=0.98$, and a weight decay of 0.05. We implement a cosine learning rate decay with a peak learning rate of 1e-4 and a linear warm-up of 5k steps. The minimum learning rate is set to 1e-5. For the LLM finetuning, following MiniGPT-4\cite{zhu2023minigpt}, we use Vicuna\cite{chiang2023vicuna} as our language decoder, constructed upon LlaMA\cite{touvron2023llama}. The total training epoch is 10 (excluding LLM finetuning), and the entire framework is implemented in PyTorch on two NVIDIA A-100 GPUs.

\subsection{Experimental Analysis}

\textbf{Comparison with multi-modal learning methods.}
Tab.~\ref{table:1} presents the performance comparison between our methods and the other multi-modal learning methods. CLIP\cite{radford2021learning} and VILT\cite{kim2021vilt}, as classic multi-modal learning methods, represent different approaches to multi-modal feature fusion. CLIP only aligns features at the feature level, requiring aggregation of image and text features before classification and detection. VILT concatenates features at the input stage but does not align image features and text features. ALBEF~\cite{li2021align} and BLIP-2 align and fuse image and text features, obtaining remarkable achievements in multi-modal downstream tasks. For a fair comparison, all the aforementioned methods perform multi-modal manipulation detection by connecting the fused features from the model's output to detection heads and classification heads and finetuning with a multi-modal manipulation detection dataset. HAMMER\cite{shao2023detecting}, proposed by the author of DGM$^4$ dataset, achieves multi-modal manipulation detection through hierarchical reasoning. The comparative results in Tab.~\ref{table:1} demonstrate that our proposed method significantly outperforms all previous methods across all evaluation metrics. This indicates the effectiveness of the locally and globally fused image feature method proposed in this paper, as well as the specially designed contrastive learning method. Specifically, compared to the current state-of-the-art algorithms, our proposed method improves accuracy by 0.9\% in binary classification, achieves a 1.7\% increase in mAP for mutil-label classification, and demonstrates a 5.5\% improvement in the F1 score for text manipulation grounding. We also draw the following observations: compared to CLIP, VILT fuses multi-modal features during the input stage, resulting in slightly superior performance across all metrics compared to CLIP. ALBEF and BLIP-2, by aligning image and text features before fusion, better leverage semantic information from both modalities, thus outperforming CLIP and VILT.

\textbf{Comparison with single-modal learning methods.}
We compared some single-modal forgery detection algorithms to verify whether the information learned from multi-modal data contributes to improving the performance of the model in a single-modal context. For the image modality, we compared our approach with two different deepfake detection methods, as shown in Tab.~\ref{table:2}. As for the text modality, shown in Tab.~\ref{table:3}, we selected the widely used large-scale pre-trained model BERT and LUKE, which demonstrated significant effectiveness in Entity Representations, for comparison. To achieve binary classification for text detection and manipulation localization, we connected the outputs of BERT and LUKE to classification heads for binary tasks and detection heads for text grounding, respectively. It can be observed from Tab.~\ref{table:2} and Tab.~\ref{table:3} that our approach significantly outperforms single-modal methods. This suggests that learning from multi-modal information contributes to enhancing the performance of single-modal tasks to some extent.

\begin{table*}[!h]
    \renewcommand\arraystretch{1.1}
    \centering
    \resizebox{0.95\textwidth}{!}{
    \begin{tabular}{cccccc|ccc|ccc|ccc}
    \toprule
    \multicolumn{6}{c|}{loss}& \multicolumn{3}{c|}{\textit{Binary Cls}} &\multicolumn{3}{c|}{\textit{Multi-Label Cls}} &\multicolumn{3}{c}{\textit{Text Grounding}}\\
    \hline
    $l_{BIC}$&$l_{MLC}$&$loss_{TMG}$&$l_{ITC}$&$l_d$&$l_v$& AUC& EER &ACC& mAP& CF1& OF1& Precision& Recall& F1\\
    \hline
    
    \checkmark & \checkmark& \checkmark&& & & 91.92& 15.71&84.65& 83.33&75.84& 77.38& 78.46& 64.15&70.59\\

    \checkmark & \checkmark& \checkmark&\checkmark& & & 93.81& 13.82&86.45& 86.63&77.57& 78.10& 75.45& 72.78&74.09\\

    \checkmark & \checkmark& \checkmark&& \checkmark& \checkmark& 93.85& 13.91&83.86& 86.65&76.38& 76.33& 72.08& 70.23&71.15\\

    \checkmark & \checkmark& \checkmark& \checkmark& \checkmark&\checkmark& \textbf{94.10}& \textbf{13.25}&\textbf{86.92}& \textbf{87.97}&\textbf{87.97}& \textbf{80.72}& \textbf{83.71}& \textbf{71.35}&\textbf{76.87}\\
    
   \bottomrule
    \end{tabular}
    }
    \caption{Ablation study of losses in the proposed method. $l_{BIC}$ represents binary classification loss, $l_{MLC}$ is the multi-label classification loss, $loss_{TMG}$ is the text manipulation detection loss, $l_{ITC}$ is the contrastive learning loss, and $l_d$, $l_v$ respectively denote the supervised losses for local and global information.}
    
    \label{table:4}
\end{table*}

\begin{table*}[!h]
    \renewcommand\arraystretch{1.1}
    \centering
    \resizebox{0.95\textwidth}{!}{
    \begin{tabular}{cc|ccc|ccc|ccc}
    \toprule
    \multicolumn{2}{c|}{Feature}& \multicolumn{3}{c|}{\textit{Binary Cls}} &\multicolumn{3}{c|}{\textit{Multi-Label Cls}} &\multicolumn{3}{c}{\textit{Text Grounding}}\\
    \hline
    global feature&local feature& AUC& EER &ACC& mAP& CF1& OF1& Precision& Recall& F1\\
    \hline
     \checkmark& & 89.96 & 18.09&82.17& 83.63&76.02& 76.66& 78.82& \textbf{73.97}&76.32\\

     & \checkmark& 81.73& 26.90&71.36& 76.25&65.21& 58.86& 58.85& 69.30&63.65\\
     
     \checkmark& \checkmark& \textbf{94.10}& \textbf{13.25}&\textbf{86.39}& \textbf{86.92}&\textbf{87.97}& \textbf{79.97}& \textbf{80.72}& 71.05&\textbf{76.87}\\
   \bottomrule
    \end{tabular}
    }
    \caption{Ablation study of global feature and local feature in the proposed method.}
    
    \label{table:5}
\end{table*}

\subsection{Ablation Study}
To further validate the effectiveness of the proposed modules in this paper, in this section, we conducted ablation experiments separately for each loss in the model and assessed the contributions of local and global features to the model.

\textbf{Ablation study of losses.}
Tab.~\ref{table:4} outlines the impact of loss functions $l_{ITC}$, $l_d$, and $l_v$ on performance. In the experiments where $l_d$ and $l_v$ were removed, we did not explicitly abandon the interaction between global and local information. Instead, the purpose was to observe whether they could still focus on different aspects of the task without any other constraints, thus enhancing the model's performance. In the ablation experiment for $l_{ITC}$, we replaced our proposed contrastive learning method with the most common contrastive learning approach from MoCo~\cite{he2020momentum}. From the table, it is evident that the removal of losses $l_d$ and $l_v$ results in a performance decline across all metrics. This indicates the effectiveness of the proposed module that fuses local and global features. Constraining local and global features with different tasks allows model to focus on differences at distinct levels, contributing significantly to the enhancement of model performance. Similarly, the removal of $l_{ITC}$ leads to a decline in model performance, signifying that our proposed contrastive learning approach effectively aligns genuine image-text pairs and distinguishes manipulated images from original images, as well as manipulated text from original text in the semantic space.

\begin{figure}
    \centering
    \includegraphics[width=1\linewidth]{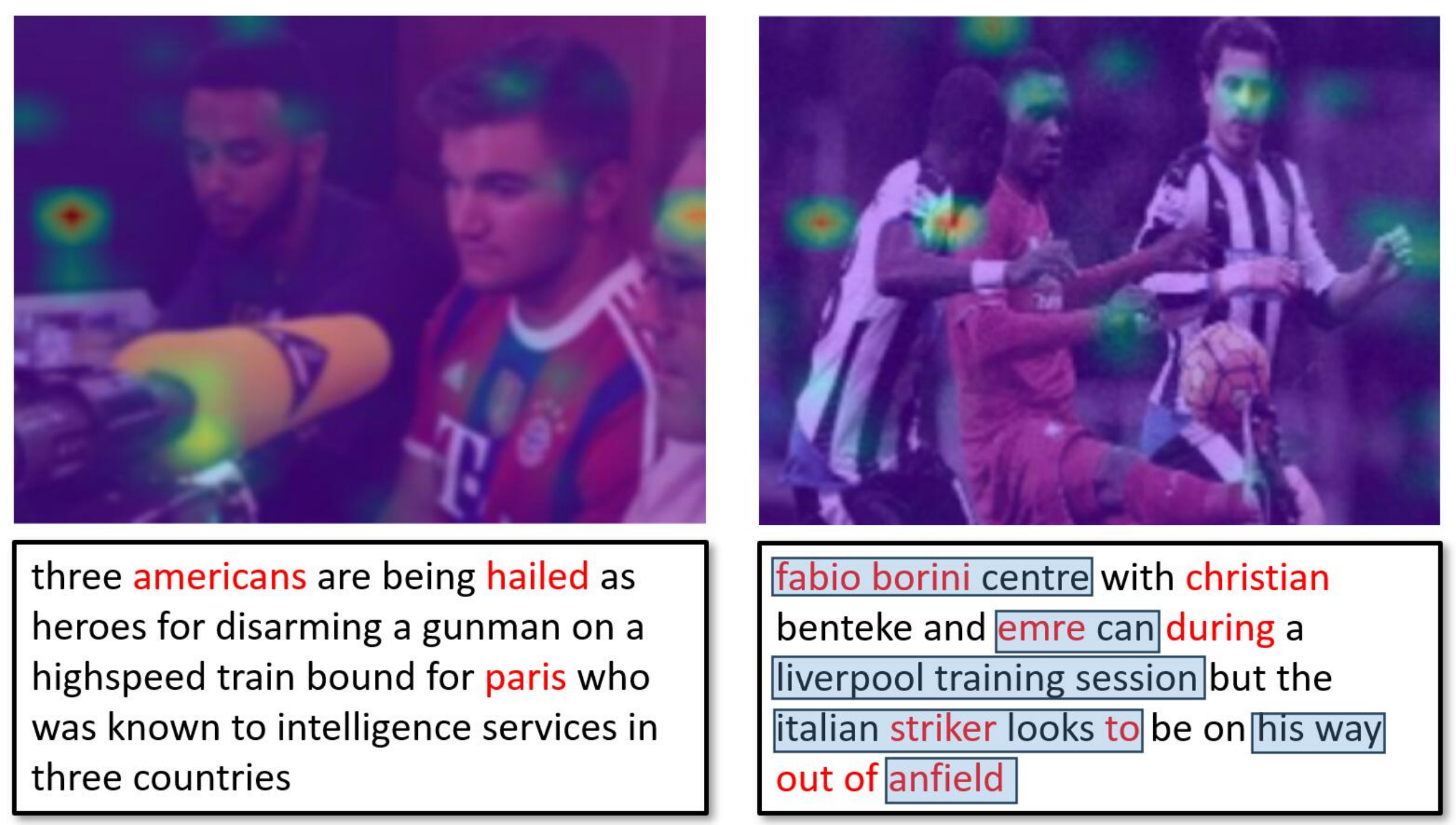}
    \caption{Visualization of attention map. The \textcolor{red}{red markings} indicate that this text carries more attention, while the \textcolor{blue}{blue boxes} denote text that has been manipulated.}
    \label{fig:attention}
\end{figure}
\textbf{Efficacy of local image feature and global feature fusion.}
In this ablation study, we removed the interaction module between global and local features, solely utilizing the features obtained from their interaction with textual information to compute the results. From Tab.~\ref{table:5}, it is evident that the absence of either global or local features has a certain impact on the results. Specifically, the lack of local features leads to a reduction in classification loss, aligning with our expectations. This is because local features, extracted using the deepfake model, contain inherent genuine information that better assists the model in distinguishing between real and fake instances and discerning manipulation types. Simultaneously, the absence of global features results in a decrease in accuracy for text manipulation localization and manipulation type classification. This is due to the lack of global semantic information in local features, making it challenging to effectively match them with the semantic information contained in the text. Consequently, the localization of forged regions in the text and the classification of text authenticity become more difficult. Clearly, in our proposed approach that both global and local features carry indispensable significance, contributing to the model's enhanced performance and understanding of multi-modal manipulation detection.

\subsection{Visualization}

To better illustrate the effectiveness of our approach in multi-modal manipulation detection, we designed visualizations for attention maps and LLM output.

\textbf{Visualization of attention map.}
We present attention map visualizations for both image and text in Fig.~\ref{fig:attention}.From the attention map, it can be observed that, when there are no forgery traces in the image, the model tends to focus more on objects that represent global semantic information in the image, such as the background and microphone in the left of Fig.~\ref{fig:attention}. However, when forgery traces are present in the image, the model shifts its attention more towards the facial region, as shown in the right of Fig.~\ref{fig:attention}. Similarly, for text information, when dealing with non-forged text, the model pays attention to vocabulary containing important semantic information, such as "Americans" and "Paris." In contrast, when dealing with text containing forgery traces, the model emphasizes semantically altered text, such as names and verb phrases.

\begin{figure*}
    \centering
    \includegraphics[width=1\linewidth]{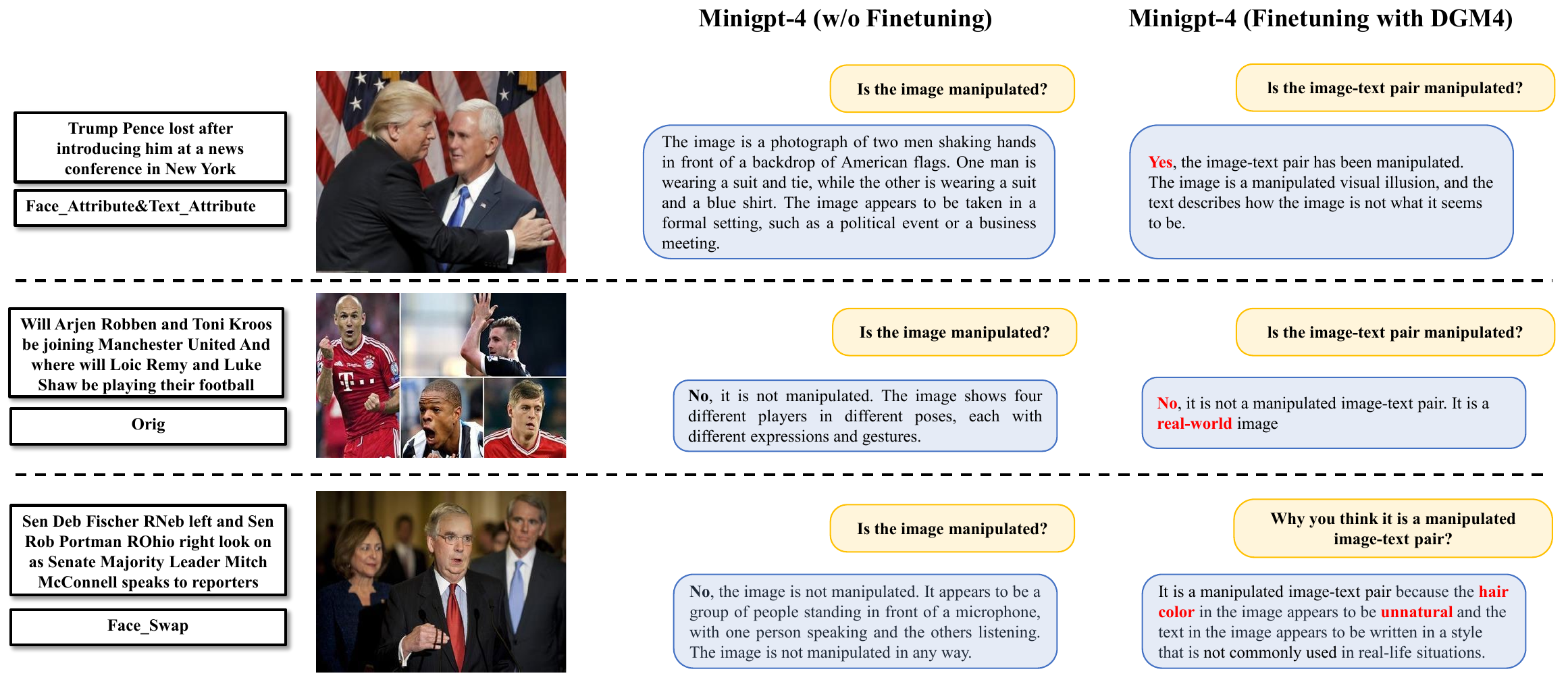}
    \caption{Visualization of LLM output. The left side displays images, corresponding text, and multi-modal manipulation labels, the right side showcases the output of the LLM without finetuning and the LLM finetuning with DGM$^4$. The yellow boxes represent questions input by humans, while the blue boxes denote responses provided by the LLM.}
    \label{fig:llm}
\end{figure*}
\label{section:LLM_Visualization}
\textbf{Visualization of LLM output.}
By employing integration of global and local features, our model effectively captures the semantic information of image-text pairs and relevant forgery-related details. Combining the model's output with the fine-tuned LLM's fully connected layers enables a more comprehensive understanding of the multi-modal manipulation detection task. This allows the LLM to assess whether an image-text pair has been manipulated based on the provided image and text, providing clear explanations. In the case of the LLM without fine-tuning, we use the image as the input and ask questions solely related to image manipulation. In contrast, for the LLM finetuning with DGM$^4$, we aggregate the image and text before inputting it into the LLM, allowing us to ask questions related to both the image manipulation and text manipulation. Fig.~\ref{fig:llm} illustrates the output results of LLM finetuning with the multi-modal manipulation detection dataset DGM$^4$, comparing them with the results without fine-tuning and showcasing the application of LLM in multi-modal manipulation detection. As illustrated in Fig.~\ref{fig:llm}, the LLM whitout finetuning often fails to provide sufficiently accurate responses when faced with questions about whether an image is manipulated, typically offering descriptions of the image content without the ability to explain the reasons for manipulation. In contrast, the LLM finetuning with DGM$^4$ can provide explicit responses and offer explanations for the reasons behind the manipulation.

\section{Conclusion}

This paper focuses on improving the performance and interpretability of the multi-modal forgery detection task. Previous research in multi-modal forgery detection has primarily concentrated on aligning features from different modalities, emphasizing global semantic information while neglecting attention to local details. Consequently, we propose M$^4$-BLIP, extracting local and global image features and text features by employing distinct encoders, aligning these features via Fine-grained Contrastive Alignment (FCA), using the Multi-modal Local-and-Global Fusion (MLGF) module to fuse these features. Finally, utilizing a feature fusion module, we merge features from different modalities and perform classification and detection on the fused features, obtaining task-relevant descriptions through the Large Language Model (LLM).

\label{sec:conclusion}

{\small
\bibliographystyle{ieeenat_fullname}
\bibliography{11_references}
}

\ifarxiv \clearpage \appendix \renewcommand\thesection{\Alph{section}}
\section{Overview}
In our supplementary material, we present the following:
\begin{itemize}
\item Details of Dataset DGM$^4$. (Section~\ref{section:Datasets})
 
\item Q-Former Implementation Details. (Section ~\ref{section:Q-Former})
 
\item Efficacy of Multi-modal Local-and-Global Fusion module. (Section ~\ref{section:Local-and-Global})
 
\item Additional Visualization of LLM. (Section ~\ref{section:LLM})
\end{itemize}

\section{Details of Dataset. }

As mentioned in Sec.~\ref{sec:experiments}, we employed the DGM$^4$ dataset as the primary dataset for our experiments. The following provides a detailed overview of the origin and composition of this dataset. 

The DGM$^4$ dataset was proposed by Shao et al.~\cite{shao2023detecting}, and the source of the original image-text pairs in the dataset is from the VisualNews dataset, comprising news from BBC, The Guardian, USA Today, and The Washington Post. Regarding the original images and text contained in the dataset, manipulation methods such as Face Swap, Face Attribute, Text Swap, Text Attribute were employed to obtain uni-modal manipulated data (i.e., either images manipulated or text manipulated exclusively). Specifically, Face Swap utilizes SimSwap~\cite{chen2020simswap} and InfoSwa~\cite{gao2021information} to manipulate faces in the images, Face Attribute employs HFGI~\cite{wang2022high} and StyleCLIP~\cite{patashnik2021styleclip} to manipulate facial attributes in the images, Text Swap uses Named Entity Recognition (NER) model to extract the person's name as a query and swaps them, Text Attribute utilizes RoBERTa~\cite{liu2019roberta} to detect the emotional sentiment in the text and swaps words containing emotional sentiments with others containing opposite emotions, thus achieving manipulation of text attributes. After obtaining uni-modal manipulated pairs, a portion of them is blended to generate mixed-manipulated pairs.

The DGM$^4$ dataset consists a total of 230k news samples, including 77,426 real image-text pairs and 152,574 manipulated pairs. The manipulated pairs contain 66,722 face swap manipulations, 56,411 face attribute manipulations, 43,546 text swap manipulations, 18,588 text attribute manipulations and 32,693 mixed-manipulation pairs.
\label{section:Datasets}

\section{Q-Former Implementation Details. }

As mentioned in Sec.~\ref{sec:Feature_Alignment_and_Fusion}, we employed the Multi-modal Local-and-Global Fusion module to extract and integrate local features, global features, and textual features. In this context, we utilized Q-Former as the model for feature extraction and fusion. The following elucidates the implementation details of Q-Former.

As illustrated in Fig.~\ref{fig:Q-Former}, after obtaining local features $e_d$ and global features $e_v$ fused features $f_d$ and $f_v$ are derived through the Q-Former, as detailed below: our Q-Former comprises three transformer blocks. The image transformer, interacting with global image features, and the image transformer interacting with local features are dedicated to extracting features from global and local visual contexts, respectively. A text transformer is employed for text encoding. Notably, two image encoders share parameters, and the text encoder shares self-attention layer parameters with the image encoder. Two learnable queries are created for input to the image transformers. Interactions with image features occur through cross-attention layers, while interactions with text involve shared self-attention layers. Simultaneously, to accommodate diverse tasks, distinct attention masks are employed. For multi-modal feature fusion, as depicted in Fig.~\ref{fig:Q-Former}, we utilize a Bi-directional Self-Attention Mask, where all queries and texts can attend to each other. The resulting query embeddings thus encapsulate multi-modal information. For contrastive learning, Uni-modal Self-Attention Mask is employed to prevent information leakage.
\begin{figure}
    \centering
    \includegraphics[width=1\linewidth]{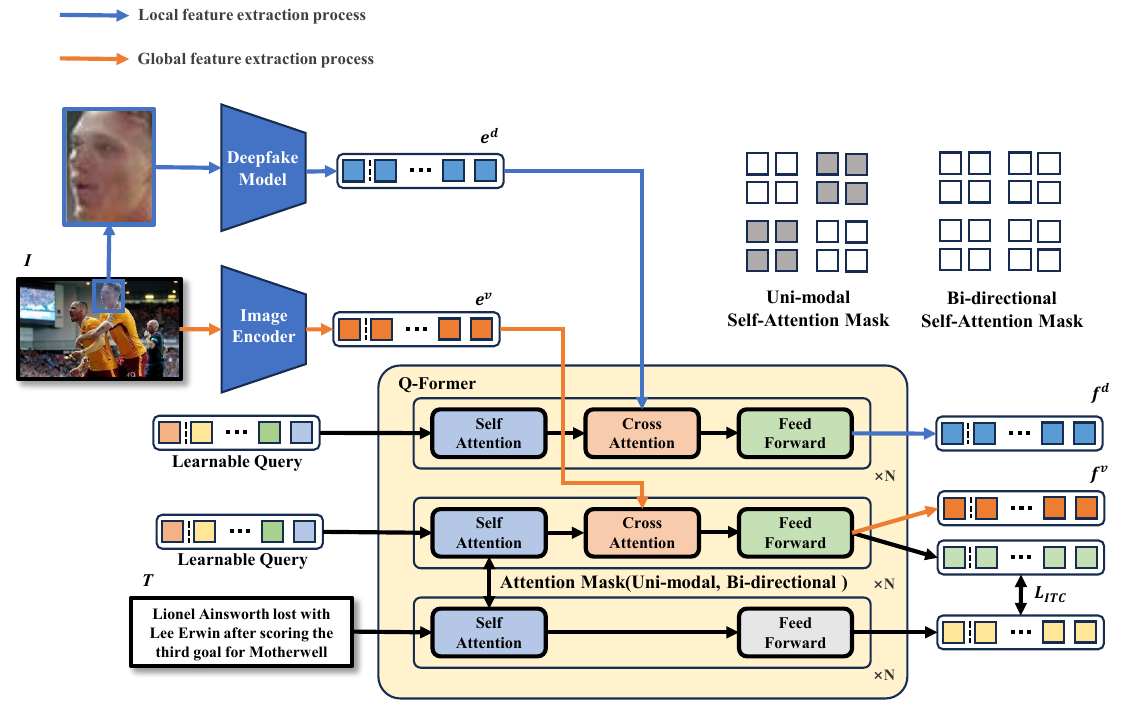}
    \caption{Implementation details of Q-Former.}
    \label{fig:Q-Former}
\end{figure}
\label{section:Q-Former}

\section{Efficacy of Multi-modal Local-and-Global Fusion module. }
\begin{table*}[!h]
    \renewcommand\arraystretch{1.1}
    \centering
    \resizebox{0.8\textwidth}{!}{
    \begin{tabular}{c|ccc|ccc|ccc}
    \toprule
    Method & \multicolumn{3}{c|}{\textit{Binary Cls}} &\multicolumn{3}{c|}{\textit{Multi-Label Cls}} &\multicolumn{3}{c}{\textit{Text Grounding}}\\
    \hline
    Categories& AUC& EER &ACC& mAP& CF1& OF1& Precision& Recall& F1\\
    \hline
    CLIP\cite{radford2021learning}& 83.22& 24.61&76.40& 66.00& 59.52& 62.31& \textbf{58.12}& 22.11& \textbf{32.03}\\

   \hline
   \textbf{CLIP+Ours}& \textbf{92.01}& \textbf{15.51}& \textbf{86.92}&\textbf{80.32}& \textbf{76.87}& \textbf{73.23}& 15.40& \textbf{27.25}&19.68\\
   \bottomrule
    \end{tabular}
    }
    \caption{The application of our local-and-global fusion method on CLIP.}
    \label{table:supp}
\end{table*}
In order to further substantiate the efficacy of the proposed local-global information fusion, we conducted experiments with our approach against common multi-modal methods, such as CLIP. Specifically, apart from obtaining image features and text features separately through the image encoder and text encoder of CLIP, we also acquired local features from facial information in the images using a deepfake model. Analogous to our Multi-modal Local-and-Global Fusion module, we initially fused global image information and local image information separately with textual information to obtain two distinct sets of features. Subsequently, we integrated these two sets of features and accomplished various tasks through different classification heads and detection heads. As depicted in Tab.~\ref{table:supp}, the incorporation of local information resulted in approximately a 10\% improvement for CLIP in both binary and multiclass classification tasks. However, there was not a substantial enhancement in the case of text manipulated localization, consistent with the findings of our ablation experiments.
\label{section:Local-and-Global}

\section{Additional Visualization of LLM. }
\begin{figure*}
    \centering
    \includegraphics[width=0.9\linewidth]{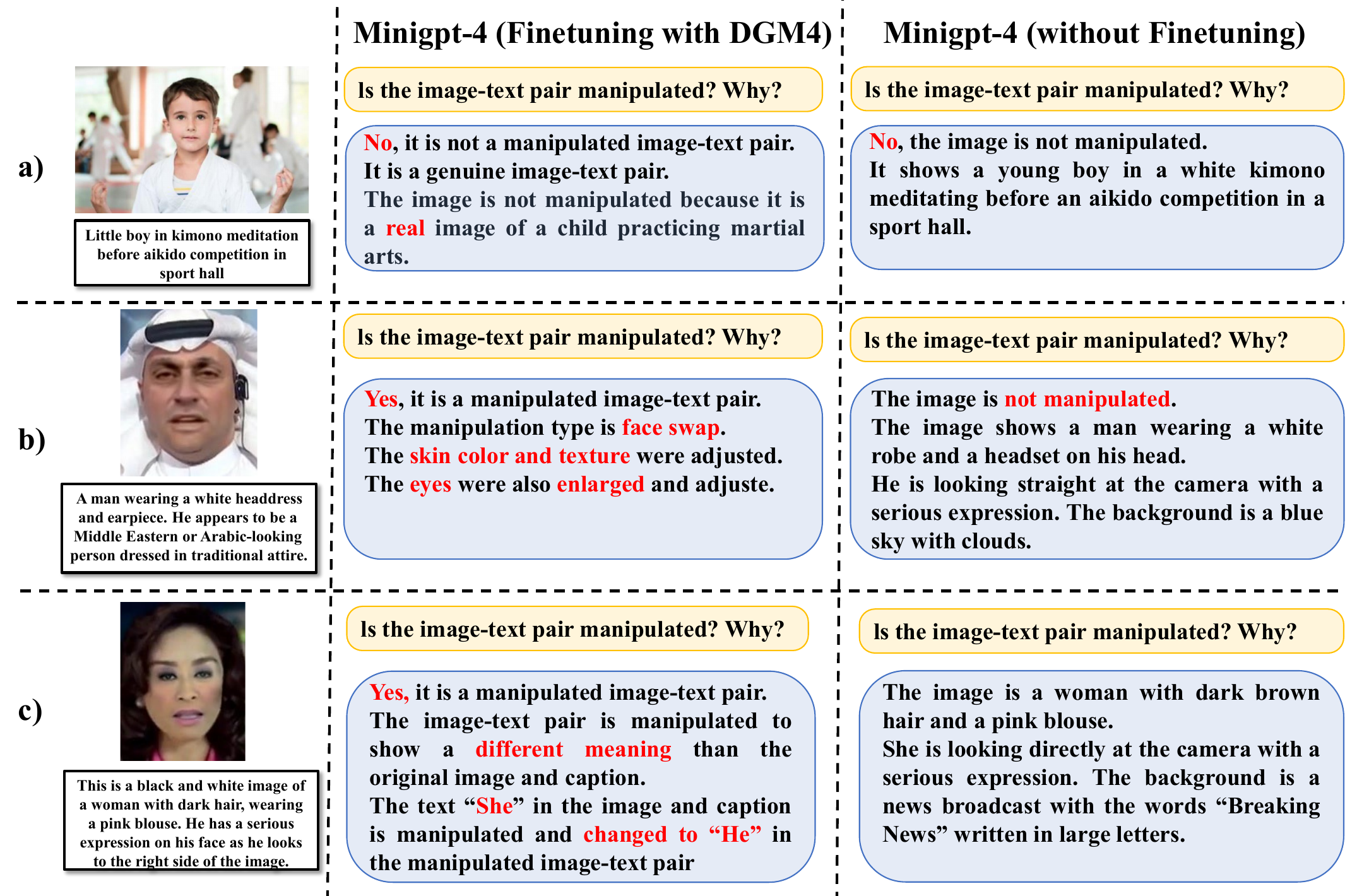}
    \caption{Additional Visualization of LLM Results. a) A real image-text pairs sourced from LAION-Face~\cite{zheng2022general}. b) and c) Two random manipulated facial images from FaceForensics++~\cite{rossler2019faceforensics++}, accompanied by corresponding descriptions generated using a multi-modal pre-trained model, aligning with the attributes of genuine facial features.}
    \label{fig:LLM-supp}
\end{figure*}
\label{section:LLM}
In order to assess the generalization capability of our model and the interpretability of the generated LLM, we conducted tests on our model using previously unseen datasets. We specifically chose real image-text pairs from the LAION-Face dataset~\cite{zheng2022general} and manipulated faces from the DeepFake dataset for testing purposes. For real image-text pairs, we directly input them into the model and inquire whether they have benn manipulated. In the case of manipulated images without text, we initially utilized a multi-modal model to generate a textual description for the synthetic image and then input the image-text pair into our model. As illustrated in Fig.~\ref{fig:LLM-supp}, compared to the MiniGPT-4 without finetuning, the fine-tuned model exhibits similar performance when confronted with real image-text pairs. However, when dealing with manipulated image-text pairs, the fine-tuned model is capable of effectively identifying whether manipulation has occurred and providing corresponding reasons. In contrast, the MiniGPT-4 without finetuning can only offer a description of the image and is unable to make judgments regarding manipulation or provide reasons.

 \fi

\end{document}